\documentclass[letterpaper, 10 pt, conference]{ieeeconf}  

\IEEEoverridecommandlockouts                              

\usepackage{times}
\usepackage{graphicx} 
\usepackage{subfigure}

\usepackage{algorithm}
\usepackage{algorithmic}

\usepackage{url}
\usepackage{amsmath}
\usepackage{graphicx}
\usepackage{verbatim}
\usepackage{stmaryrd}
\usepackage{url}

\DeclareMathOperator*{\argmin}{arg\,min}

\long\def\ignorethis#1{}

\newcommand{\inv}{^{-1}}

\newcommand{\gauss}{\mathcal{N}}

\newcommand{\trace}{\text{tr}}

\renewcommand{\eqref}[1]{Equation~(\ref{#1})}

\newcommand{\dataset}{\mathcal{D}}

\newcommand{\trajdist}{q}
\newcommand{\policy}{\pi}

\newcommand{\return}{J}
\newcommand{\params}{\theta}

\newcommand{\cost}{c}
\newcommand{\state}{\mathbf{x}}
\newcommand{\action}{\mathbf{u}}

\newcommand{\traj}{\tau}

\newcommand{\trajmu}{\mu^\trajdist}

\newcommand{\polmu}{\mu^\policy}
\newcommand{\polsig}{\Sigma^\policy}

\newcommand{\ucovar}{\mathbf{C}}

\newcommand{\kl}{D_\text{KL}}

\newcommand{\st}{\state_t}
\newcommand{\at}{\action_t}

\title{\LARGE \bf
Reset-Free Guided Policy Search: Efficient Deep Reinforcement Learning with Stochastic Initial States
}

\author{William Montgomery$^{\star,1}$, Anurag Ajay$^{\star,2}$, Chelsea Finn$^{2}$, Pieter Abbeel$^{2}$, and Sergey Levine$^{2}$
\thanks{$^{\star}$The first two authors contributed equally to this work}%
\thanks{$^{1}$Department of Computer Science and Engineering, University of Washington, Seattle, WA 98195}%
\thanks{$^{2}$Department of Electrical Engineering and Computer Sciences, University of California, Berkeley, Berkeley, CA 94709}%
}

\begin{document}

\maketitle
\thispagestyle{empty}
\pagestyle{empty}

\begin{abstract}
Autonomous learning of robotic skills can allow general-purpose robots to learn wide behavioral repertoires without extensive manual engineering. However, robotic skill learning must typically make trade-offs to enable practical real-world learning, such as requiring manually designed policy or value function representations, initialization from human demonstrations, instrumentation of the training environment, or extremely long training times. We propose a new reinforcement learning algorithm that can train general-purpose neural network policies with minimal human engineering, while still allowing for fast, efficient learning in stochastic environments. We build on the guided policy search (GPS) algorithm, which transforms the reinforcement learning problem into supervised learning from a computational teacher (without human demonstrations). In contrast to prior GPS methods, which require a consistent set of initial states to which the system must be reset after each episode, our approach can handle random initial states, allowing it to be used even when deterministic resets are impossible. We compare our method to existing policy search algorithms in simulation, showing that it can train high-dimensional neural network policies with the same sample efficiency as prior GPS methods, and can learn policies directly from image pixels. We also present real-world robot results that show that our method can learn manipulation policies with visual features and random initial states.
\end{abstract}

\section{Introduction}

General-purpose robots operating in the real world will require large repertoires of flexible and robust motion skills, particularly in the domain of robotic manipulation.
Reinforcement learning (RL) algorithms aim to autonomously learn such skills, to avoid the need for extensive hand-engineering that scales poorly as the size of the motor skill repertoires grows.
While RL methods have been applied widely to a range of tasks~\cite{pma-reps-10, tbs-rlmsh-10}, practical real-world applications often rely on simplifying assumptions to achieve tractable learning, including hand-designed policy or value function representations~\cite{ins-lallm-03}, demonstrations by experts~\cite{phas-lgmsl-09}, or training in an instrumented environment~\cite{lfda-eetdv-16}.
Continuous deep RL methods, which use general-purpose neural network function approximators~\cite{slmja-trpo-15,lhphe-ccdrl-16}, avoid the need for many of these assumptions, but suffer from extremely poor sample efficiency due to their high-dimensional parameter space, precluding their use on real robotic systems.

Guided policy search (GPS) methods~\cite{lfda-eetdv-16} mitigate the issue of sample efficiency in deep RL by restructuring the reinforcement learning objective into two distinct phases: a simple trajectory-centric RL phase (the ``control'' phase, or C-phase) that only optimizes trajectories from a small set of consistent initial states (thereby simplifying the problem), and a supervised learning phase (the S-phase). This can be viewed as a type of dual decomposition~\cite{b-co-04}, where the complex original RL problem is broken up into multiple simpler ones. The trajectory-centric RL in the C-phase acts as a ``teacher,'' generating training data for the S-phase. The S-phase in turn can train a general-purpose neural network policy using standard, scalable supervised learning methods, which accounts for the efficiency of this approach.

One of the major limitations of GPS is the need to decompose the problem during the C-phase, which is typically done by assuming that the environment can be deterministically reset into one of $M$ initial states, and that the task can be performed episodically from each of these $M$ states using a separate trajectory-centric \emph{local policy}, optimized with a scalable, but comparatively weak trajectory-centric RL algorithm.
This is typically a model-based algorithm similar to iterative LQR~\cite{tl-gilqg-05}. It is precisely this decomposition of the problem into $M$ initial states that allows for trajectory-centric RL to be used to supervise the neural network policy, but this assumption is also very restrictive. Not all RL problems allow for such deterministic resets: arbitrary stochastic environments might choose a random initial state for each trial beyond the agent's control. Furthermore, restricting learning to a small fixed set of initial states limits the generalization of the resulting global policy. In this work, we extend guided policy search to relax this assumption, while still enabling efficient trajectory-centric RL to be used for supervision.

The main contribution of our paper is a novel guided policy search (GPS) algorithm that can train high-dimensional nonlinear policies, such as deep neural networks, for arbitrary stochastic environments.
In order to remove the deterministic reset assumption and enable the method to be used with fully stochastic initial states, we draw samples directly from the global neural network policy, rather than the local policies, as is standard in most prior work, and then cluster the initial states to instantiate new local policies on the fly at each iteration.
These new local policies are then improved as usual in the C-phase, using efficient trajectory-centric RL.
Our method preserves the sample efficiency of GPS while making it applicable to a more general set of stochastic RL problems. This not only improves the applicability of the algorithm, but also allows us to train policies with superior generalization, since policies which experience a wide range of randomized initial states will generalize better to novel states at test time. Our experimental evaluation confirms this hypothesis through comparisons on a set of challenging simulated and real-world manipulation tasks. We also compare our approach to state-of-the-art deep reinforcement learning methods, in terms of both sample efficiency and performance, demonstrating that our approach substantially outperforms prior GPS algorithms on generalization and achieves success rates comparable to prior direct deep RL methods with one to two orders of magnitude fewer samples. We also show that our method can learn policies from raw image pixels, represented by convolutional neural networks, and can learn a real-world robotic pushing skill with visual features and completely random manual resets.

\section{Related Work}

Robotic reinforcement learning (RL) sets out to tackle the very general question of how new behaviors can be learned automatically from only high-level cost or reward specifications~\cite{sb-irl-98,tbs-rlmsh-10,pma-reps-10,kbp-rlrs-13,dnp-spsr-13}. However, the high dimensionality of robotic systems, high variability of unstructured open-world environments, and the practical constraints of learning with real physical hardware typically demand additional restrictions on the highly general reinforcement learning formulation to enable effective applications in the real world. One of the most common and effective methods for taming the complexity of real-world robotic learning is to provide the algorithm with hand-engineered representations that are well suited to the task at hand. Such representations can range from very specific policy classes for flight~\cite{acqn-arlah-06} or locomotion~\cite{tzs-spgrl-04} to general-purpose dynamic movement primitives~\cite{ins-lallm-03}. However, these representations require considerable expertise to design, and typically constrain the policy in some way, for example by only allowing it to encode feedbacks around a single trajectory. Practical real-world RL also often makes use of example demonstrations to bootstrap learning~\cite{krps-lfcpc-11}.

Recently, continuous deep reinforcement learning methods~\cite{slmja-trpo-15,lhphe-ccdrl-16} have been shown to solve complex tasks using general-purpose deep neural network representations. These methods greatly reduce the burden of manual policy class and feature engineering, but typically impose prohibitive sample requirements. For example, the TRPO algorithm~\cite{slmja-trpo-15} requires about 200,000 episodes to learn a bipedal locomotion gait, the equivalent of about 20 days of real time experience.

An alternative to direct deep RL algorithms based on temporal difference error and policy gradients is to use guided policy search (GPS) methods~\cite{lfda-eetdv-16,ml-gpsamd-16}. These algorithms transform the policy search problem into a supervised learning problem, with supervision provided by simple trajectory-centric RL methods. These trajectory-centric methods resemble the trajectory-centric approaches popular in robotics~\cite{tbs-rlmsh-10,pma-reps-10}, but instead of training the policy directly, they simply provide supervision for training a nonlinear deep neural network policy from multiple different instances of the task (e.g. different poses of a target object). During the supervised training phase, the input to the policy can be changed, for example to exclude ground truth object positions and instead include camera images, thus training policies that combine perception and control~\cite{lfda-eetdv-16}. The main limitation of such methods is that they require the ability to reset the environment deterministically into each of the initial states for each of the trajectory-centric controllers. In this work, we lift this limitation to enable the application of GPS methods to general reinforcement learning problems.

\section{Background and Preliminaries}
We consider stochastic finite-horizon problems, with states $\st$ and actions $\at$, which may be characterized by an unknown dynamics distribution $p(\state_{t+1} | \st, \at)$, a state-action cost $\cost(\state, \action)$, and a horizon length $T$.
For convenience, we use $\traj$ to refer to the entire trajectory sequence $(\state_1, \action_1, \dots, \state_T, \action_T)$.
We aim to learn a policy $\policy_\params(\action|\state)$, parameterized by $\params$, which minimizes the expected total cost:
\begin{align*}
   \params \leftarrow \min_{\params} \sum_{t=1}^T E_{\policy_\params(\st, \at)} [\cost(\st, \at)],
\end{align*}
where we use $\policy_\params(\st, \at)$ to refer to the time-indexed marginals of $\policy_\params(\traj) = p(\state_1) \prod_{t=1}^T p(\state_{t+1}|\st, \at) \policy_\params(\at|\st)$. For convenience we will use $\return(\params)$ to refer to expected total cost $\sum_{t=1}^T E_{\policy_\params(\st, \at)} [\cost(\st, \at)]$.

Policy gradient methods work by approximating the gradient of the expected cost with respect to $\params$ using rollouts, but this becomes sample inefficient when $\params$ is high-dimensional due to the high variance of the sample-based gradient estimator~\cite{dnp-spsr-13,slmja-trpo-15}. Guided policy search methods simplify the optimization by splitting it into a trajectory-centric RL phase (C-phase) and a supervised learning phase (S-phase). During the C-phase, the trajectory-centric RL method uses the most recent samples to produce a feedback controller for each of the $M$ initial positions. We refer to these controllers as local policies. In the S-phase, the global policy is trained using standard supervised learning to match the output of each of the local policies. Training high-dimensional function approximators is substantially easier and more stable using supervised learning as compared to policy gradient methods, which are difficult to scale beyond about 100 parameters~\cite{dnp-spsr-13} without using extremely large numbers of samples~\cite{slmja-trpo-15}.

At convergence, the local policies have been optimized to minimize the cost, while the global policy has been trained to mimic the local policies. In general, not all time-varying local policies can be mimicked by a single global policy, so GPS takes additional steps to ensure that the local policies are close to the space of policies that can be represented by the global policy class. In the mirror descent guided policy search (MDGPS) algorithm, which we extend in this work, this is accomplished by constraining the local policies to stay close to the global policy, in terms of their KL-divergence. Prior GPS methods use other types of constraints, including soft constraints~\cite{lfda-eetdv-16}.

\begin{algorithm}
    \caption{MDGPS \label{alg:mdgps}}
	\begin{algorithmic}[1]
		\FOR{iteration $i \in \{1, \dots, I\}$}
        \STATE Generate datasets $\{\dataset_m\}$ by running $\policy_{\params}$ or $\trajdist_m$ $N$ times from each initial position $\state_1^m$ for $m \in \{1, \dots, M\}$
		\STATE Adjust $\epsilon$ (see Appendix~\ref{sec:step})  
        \STATE C-phase: Update each local policy $\trajdist_m$ using trajectory-centric RL on $\dataset_m$
        \STATE S-phase: Update global policy $\policy_\params$ to match the local policies using supervised learning
		\ENDFOR
	\end{algorithmic}
\end{algorithm}

The MDGPS optimization, illustrated in Algorithm~\ref{alg:mdgps}, repeatedly performs the C-phase and S-phase updates to improve the policy. As implied by the name, this can be interpreted as an approximate form of mirror-descent on $\return(\params)$~\cite{ml-gpsamd-16, bt-mdnps-03}, constrained to the set of policies, $\Pi_\params$, which are represented by the function approximator (e.g., a large neural network). In the C-phase, we learn new local policies for each initial position, and in the S-phase we ``project'' these down to a single policy in $\Pi_\params$, using KL divergence as the distance metric. Below we describe the two phases of the algorithm in more detail.

\subsection{Trajectory-Centric Reinforcement Learning}
\label{sec:bgtraj}

The C-phase involves optimizing each of the local policies $\trajdist_m$ with respect to the cost, constrained to stay within a fixed KL-divergence of the global policy $\policy_\params$. In order to make the optimization feasible, MDGPS makes several approximations and assumptions. The first assumption is that each local policy can be represented by a time-varying linear-Gaussian controller, $\trajdist_m(\st|\at) \sim \gauss(K_{m,t} \state_t + k_{m,t} , \ucovar_{m,t})$. This is reasonable for systems where stochasticity is low or Gaussian, and most of the randomness comes from the initial state distribution (which is fixed to a single sample for each local policy). Second, MDGPS uses linear regression to fit time-varying linear-Gaussian dynamics $p_m(\state_{t+1}|\st, \at) = \gauss{(f_{xt} \st + f_{ut} \at + f_{ct} , F_t)}$ to the $N$ samples drawn from that initial position. To reduce the number of samples needed for linear regression, a GMM prior on tuples $(\st, \at, \state_{t+1})$ may be used, as in prior work~\cite{la-lnnpg-14}. Finally, in order to handle the constraint against the global policy, MDGPS uses a time-varying linear-Gaussian approximation to the global policy, $\bar{\policy}_m (\at|\st) = \gauss{(K_{m,t}\st + k_{m,t} , \Sigma^{\policy}_{m,t})}$. This linearization is also constructed using linear regression, with a GMM prior on tuples $(\st, \at)$. The constrained optimization is then
\begin{align*}
\trajdist_m \leftarrow \argmin_\trajdist E_{\trajdist(\traj)}\left[\sum_{t=1}^T \cost(\st,\at) \right] \text{ s. t. } \kl(\trajdist||\bar{\policy}_m) \leq \epsilon,
\end{align*}
where $\epsilon$ is the constraint on the KL divergence. As shown in previous work~\cite{la-lnnpg-14}, we can optimize this with a dual gradient descent method, optimizing the primal variables with an LQR backward pass (after approximating the cost function with a quadratic Taylor expansion), and optimizing the single dual variable with gradient descent or a linesearch. This provides an efficient and simple method for model-based local policy improvement, while respecting the KL-divergence constraint.

\subsection{Supervised Learning}
Given the optimized local policies, ${\trajdist_m}$, we can train the global policy with supervised learning. MDGPS uses a KL divergence objective of the form
\begin{align*}
\min_\params \sum_{m=1}^M \sum_{t=1}^T E_{\trajdist_m(\state_{t,m})} \left[  \kl(\policy_\params(\at|\state_{t, m}) || \trajdist_m(\at|\state_{t, m})) \right],
\end{align*}
which is approximated with samples to obtain
\begin{align*}
\min_\params \sum_{m=1}^M \sum_{t=1}^T \sum_{n=1}^N \kl(\policy_\params(\at|\state_{t, m, n}) || \trajdist_m(\at|\state_{t, m, n})),
\end{align*}
where $\state_{t, m, n}$ refers to the state at time $t$ of the $n^\text{th}$ sample of the $m^\text{th}$ condition. When using a conditionally Gaussian policy, $\policy_\params(\at|\st) = \gauss(\mu^{\policy}(\st), \Sigma^{\policy}(\st))$, this objective has a convenient form that resembles least squares:
\begin{align*}
&\min_\theta \sum_{t,m,n} \trace[\ucovar_{tm}\inv\polsig(\state_{t,m,n})] - \log|\polsig(\state_{t,m,n})| + \\
&(\polmu(\state_{t,m,n}) - \trajmu(\state_{t,m,n}))\ucovar_{tm}\inv(\polmu(\state_{t,m,n}) - \trajmu(\state_{t,m,n})),
\end{align*}
where we use $\trajmu(\state_{t,m,n})$ to refer to the mean of the local policy. When the policy covariance is independent of state, this objective is simply a weighted Euclidean loss on the mean $\polmu(\state_{t,m,n})$, which might be parameterized, for instance, by a deep neural network~\cite{lfda-eetdv-16}.

\subsection{Sampling}

The samples on line 2 of Algorithm~\ref{alg:mdgps} in MDGPS can in principle be generated from any distribution, so long as they are useful for estimating the local dynamics around each $\trajdist_m$. Prior GPS algorithms typically generate the samples from $\trajdist_m$ directly~\cite{lfda-eetdv-16}, since these samples are most likely to be in the regions where accurate dynamics estimates are required. However, using MDGPS it is also possible to generate the samples from the global $\policy_\params$, which we refer to as on-policy samples. This is a reasonable choice because the local policy updates are constrained to stay close to $\policy_\params$, and therefore samples from $\policy_\params$ are likely to also be in the regions where accurate dynamics estimates are required. However, prior MDGPS methods still require the ability to reset the environment into each of the $\state_1^m$ initial states during training, so as to generate $N$ samples from each state for dynamics estimation, regardless of which policy is actually used for taking the samples. In the next section we describe our method, which extends MDGPS to the case of stochastic initial states, where the algorithm does not assume any ability to control the initial state $\state_1$.

\section{Reset-Free Guided Policy Search with Stochastic Initial States}

\begin{algorithm*}[t]
    \caption{MDGPS with Trajectory Aware Clustering \label{alg:mdgps_cluster}}
	\begin{algorithmic}[1]
		\FOR{iteration $i \in \{1, \dots, I\}$}
		\STATE Generate $M$ sample trajectories ${\traj}_m$ by running $\policy_{\params}$ from randomly selected initial state $\state_1^m$
        \STATE Randomly assign trajectories to each of $K$ clusters, producing datasets $\{\dataset_k\}$
        \WHILE{cluster assignments not converged}
            \STATE Fit each linear-Gaussian dynamics 
$p_k(\state_{t+1}|\st,\at)$ using samples in each $\dataset_k$
            \STATE Fit each linearized global policy $\bar{\policy}_k(\at|\st)$ using samples in each $\dataset_k$
            \STATE Compute probabilities of each sample being drawn from each trajectory cluster (Eq.~\ref{eq:estep})
            \STATE Assign trajectories to cluster with highest probability, producing new datasets ${\dataset_k}$
        \ENDWHILE
		\STATE Adjust $\epsilon$ (see Section~\ref{sec:step})
        \FOR{sample $m \in \{1,\dots,M\}$, in parallel}
        \STATE C-phase: Compute per-sample local policies $\trajdist_m$, using dynamics and global policy linearizations of the corresponding cluster and per-sample cost expansion around $\traj_m$ (see Section~\ref{sec:multilocal})
        \ENDFOR
        \STATE S-phase: Update global policy $\policy_\params$ to match new local policies using supervised learning
		\ENDFOR
	\end{algorithmic}
\end{algorithm*}

The guided policy search method described in the previous section relies on deterministic resets to enable the use of highly efficiency trajectory-centric RL based on LQR: each local policy can only succeed from a particular initial state $\state_1^i$, and in order to train each policy, the robot must be able to repeatedly reset the environment into each $\state_1^i$ on demand. In principle, the algorithm converges to a locally optimal global policy if the set $\{\state_1^i\}$ is sampled from the underlying initial state distribution $p(\state_1)$. However, because the set of initial states $\{\state_1^i\}$ is kept fixed during training, we might need a large number of these states to adequately cover a broad initial state distribution. Furthermore, the assumption that the robot can always reset the environment into a chosen initial state $\state_1^i$ is not always practical, and might require additional manual engineering to design an effective automated reset system. In this section, we discuss how the MDGPS algorithm described in the previous section can be extended to handle completely random initial states drawn from $p(\state_1)$. Since this will remove the deterministic reset assumption, it will make MDGPS broadly applicable to the general reinforcement learning problem setting.

An overview of our method is provided in Algorithm~\ref{alg:mdgps_cluster}.
The structure of the algorithm resembles the on-policy variant of MDGPS described in the previous section, with the addition of a dynamic assignment of samples to local policies. Since each sample now has a different random initial state, we cannot directly match the samples with corresponding local policies. However, we can cluster the samples at each iteration and dynamically instantiate a local policies for each cluster. Because sampling is performed on-policy using the global policy $\policy_\params(\at|\st)$, the local policies are never actually executed, and are only instantiated within each iteration to construct supervision for updating the behavior of the global policy in the S-phase. In practice, we actually instantiate multiple local policies within each cluster, as discussed in Section~\ref{sec:multilocal}. In the following subsections, we discuss the particular clustering approach that we use and provide a walkthrough of the complete method in Algorithm~\ref{alg:mdgps_cluster}.

\subsection{Trajectory-Aware Clustering}
When fitting dynamics and policy linearizations, we are essentially assuming that each initial position corresponds to a single trajectory distribution, defined by $p_m(\state_{t+1}|\st, \at)$ and $\bar{\policy}_m(\at|\st)$, which are fit using samples from that initial position. When using deterministic resets, as in standard GPS, it makes sense to fit a single trajectory distribution to each initial condition, but in theory we can fit a linearization to any set of samples. The linearizations are most useful for trajectory optimization when the disributions fit ``tightly'' around their samples, and so we propose an expectation-maximization approach for clustering samples and fitting their linearizations.

For each cluster, we want to fit time-varying models $\bar{\policy}_k(\at|\st)$ and $p_k(\state_{t+1}|\st, \at)$, just as we did previously for each initial position, as well as an initial state probability $p_k(\state_1)$ since we no longer have consistent initial positions. We use a Gaussian distribution to parametrize $p_k(\state_1)$. Now, since $\bar{\policy}_k$ and $p_k(\state_{t+1}|\st, \at)$ are both linear-Gaussian, and the full trajectory distribution corresponding to cluster $k$ is given by
\[
p_k(\traj) = p_k(\state_1) \prod_{t=1}^T \bar{\policy}_k(\at|\st)p_k(\state_{t+1}|\st, \at),
\]
we can see that the trajectory distribution is itself Gaussian. For convenience, we will refer to all of the parameters of this distribution together as $\phi_k$. These parameters include the linearization of the global policy, the linearization of the dynamics, and the initial state Gaussian.

Given this model, we perform expectation-maximization using a mixture of trajectory distributions, where we use $P(k)$ to refer to the mass on each mixture element.
In the E-step, we compute the probability of each sample being drawn from each trajectory:
\begin{align}
    p(\traj_m | \phi_k) \propto P(k) p_k(\traj_m).\label{eq:estep}
\end{align}
In the M-step, we assign each trajectory to its most likely cluster, and compute the new parameters $\phi_k$ using the samples in that cluster, using the linearization approach described in Section~\ref{sec:bgtraj}. For computational efficiency, we use hard EM rather than soft cluster assignments. We found that the E-step almost always produced probabilities near 0 or 1, which is likely a consequence of the high dimensionality of the trajectory distribution, making hard EM updates a reasonable and highly efficient approximation.

\begin{figure*}[t]
    \centering
    \includegraphics[width=0.99\textwidth]{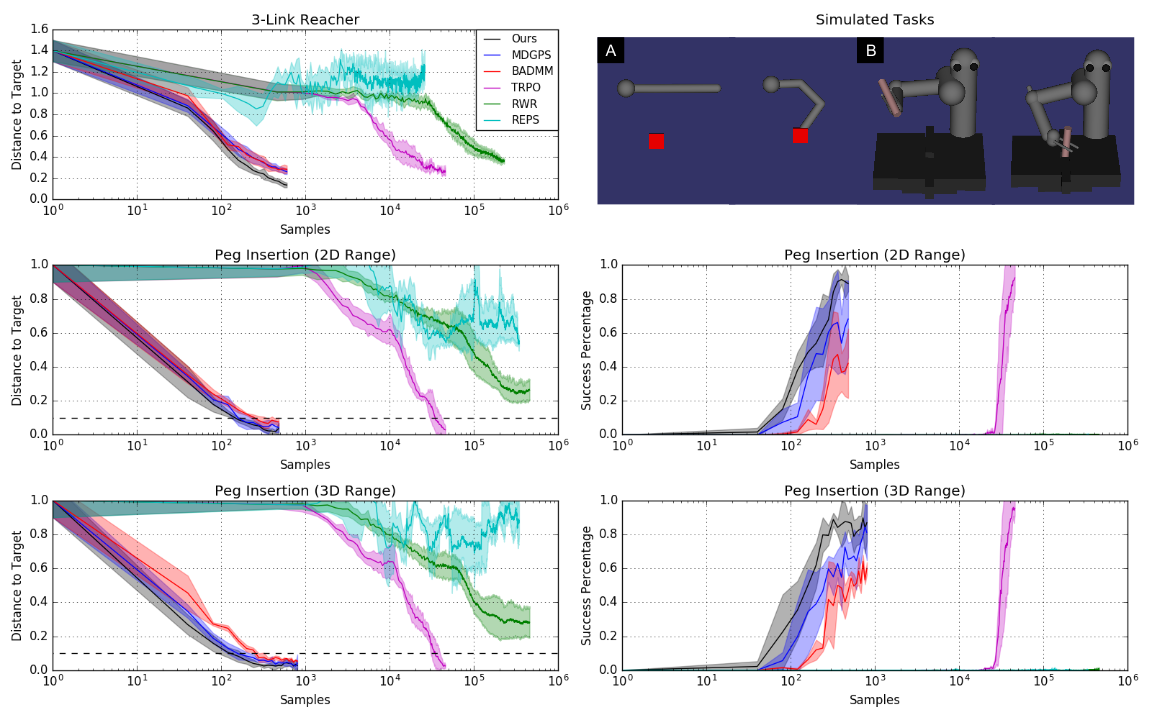}
    \vspace{-0.1in}
    \caption{Simulated comparisons. We compare our method to prior deep RL methods on several simulated tasks, using deterministic resets only for prior GPS methods. (A) A 3 DoF arm must learn to reach a random target position with its end-effector. (B) A 7 DoF arm must learn to insert a peg into the hole on a randomly placed table. On both reacher and peg insertion tasks, the final distance between the end effector and target is shown vs. the number of samples used during training. For peg insertion, the height of the hole is shown as a dotted line, and we also include the percentage of test trials which successfully place the peg into the hole. Our method maintains the sample efficiency of prior GPS methods, while relaxing the assumption that we can deterministically reset to consistent initial positions during training. Prior GPS methods are allowed to use deterministic resets into consistent initial states in these experiments. Note that the samples are on a log scale, to make it possible to illustrate the prior RL algorithms.}
    \vspace{-0.1in}
    \label{fig:sim_results}
\end{figure*}

\subsection{Instantiating Local Policies}
\label{sec:multilocal}

During trajectory-aware clustering, we fit $K$ time-varying models for the dynamics and policy linearizations, which are then fed into the C-phase. We could simply instantiate one local policy for each cluster and update it in the same way as in standard GPS, using the dynamics corresponding to that cluster. We found that this generally produced reasonable results, but could be improved with a simple modification. Note that the C-phase also requires a quadratic expansion of the cost function for the LQR backward pass. When using deterministic resets to consistent initial conditions, as with standard GPS methods, the quadratic expansion can be performed over the mean of each trajectory distribution, or averaged over the $N$ samples originating from an initial state. If the cost function is globally quadratic, this produces the same quadratic cost approximation. However, for complex non-quadratic functions, choosing the right trajectory around which to expand the cost can be very important. In our method, since the samples in a single cluster typically do not even originate from the same initial state, this becomes even more important.

We found that we could obtain substantially better performance if we avoid this approximation, and instead expand the cost function around each of the samples, performing a separate LQR backward pass for each one. This results in not $K$ but $M$ separate local policies at each iteration, one for every sample. Although this increases the computational cost, it substantially improves performance in the presence of non-quadratic cost functions. Furthermore, since all of the local policy updates are independent of one another, they can be trivially parallelized. In practice, we found that the cost of the local policy updates was negligible when compared to the cost of fitting dynamics, clustering the trajectories, and training the global policy with supervised learning.

\begin{figure*}[t]
    \centering
    \includegraphics[width=0.99\textwidth]{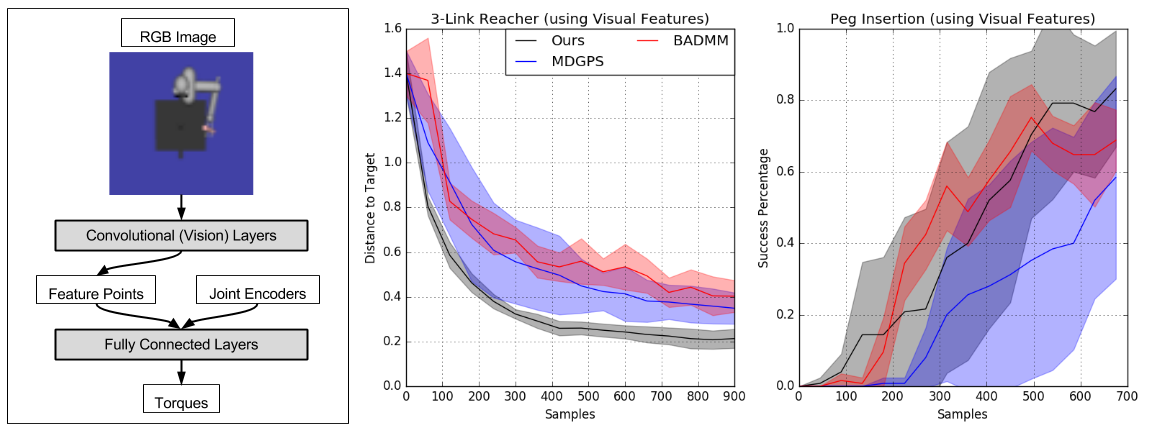}
    \vspace{-0.1in}
    \caption{Learning from raw pixels. The diagram (left) shows the network architecture: RGB images are passed into a convolutional network, which produces spatial feature points. These features are then fed into fully connected layers, together with the joint angles/velocities. We compare our method to prior GPS methods.}
    \vspace{-0.1in}
    \label{fig:vis_results}
\end{figure*}

\subsection{Global Step Size Adjustment}
\label{sec:step}
The local policy step size $\epsilon$ plays an important role in the performance of GPS methods. Since we are using time-varying linear approximations to the unknown (possibly nonlinear) dynamics, large changes to the local policies caused by large values of $\epsilon$ might overestimate the improvement in cost, resulting in unstable behavior and divergence. On the other hand, small values of $\epsilon$ slow down learning and can increase the sample complexity of the method. Prior GPS methods set $\epsilon$ individually for each local policy by comparing the expected and actual improvement from the previous iteration~\cite{lfda-eetdv-16,ml-gpsamd-16}, since each policy uses a separate and independent dynamics linearization. Since our method dynamically changes the assignment of local policies to initial states, there is no notion of a single previous local policy for every local policy at the current iteration, and we therefore use a single value of $\epsilon$ for all local policies. This value is updated in the same way, by comparing the actual and expected improvement from the previous iteration, but averaged over all local policies.

Formally, let $\Delta \return$ be the actual improvement in the cost, given by the difference in the cost from the previous iteration, and let $\bar{\Delta} \return$ be the expected improvement, given by the C-phase on the previous iteration. Both of these quantities can be computed analytically under the current and previous linear-Gaussian dynamics and quadratic cost, via the formula for the expectation of a quadratic form under a Gaussian~\cite{ml-gpsamd-16}. In our method, these expectations are averaged over the local policies $q_1,\dots,q_M$. As in prior work, the updated step size $\epsilon^\prime$ is given by $\epsilon^\prime = \epsilon(\bar{\Delta} \return)/2(\bar{\Delta} \return - \Delta \return)$.

\subsection{Algorithm Summary}
We can now summarize the entire algorithm, given in Algorithm~\ref{alg:mdgps_cluster}. On line 2, we sample $M$ trajectories by executing the current global policy. Since we are now assuming that the environment is stochastic, each sample begins at a different initial position, sampled from the unknown distribution $p(\state_1)$. On line 3, we initialize the EM procedure by randomly assigning trajectories to each of datasets $\mathcal{D}_k$ and performing an initial M-step. Next, we repeat the M and E-steps, fitting $\phi_k$ for each cluster and then reassigning the samples until the assignments have converged. On line 10 we adjust $\epsilon$ using the step size rule previously discussed. Finally, on lines 12 and 14, we perform the usual C-phase and S-phase updates, producing new local policies and using them for supervised learning of the global policy.

\section{Experimental Evaluation}

Our experimental evaluation aims to answer the following questions: (1) Can our proposed GPS algorithm learn high-dimensional robotic tasks with random initial states? (2) Can the use of random initial states improve the generalization of the algorithm on tasks where the initial states have a wide distribution, as compared to the consistent initial states in prior GPS methods? (3) How does our algorithm compare to other deep reinforcement learning algorithms that similarly can handle random initial states? (4) Can our method learn policies which combine perception and control? (5) Can our method be used to learn policies on real physical robots with random initial states? We address these questions with both simulated and real-world experiments, discussed below.

\subsection{Simulated Robotic Policy Learning}

Our simulation environment is based on the MuJoCo simulator~\cite{tet-mjc-12}, which we use to simulate two robotic manipulation tasks: a planar 2D reaching task with a 3 DoF arm, and a more complex peg insertion task with a simulated PR2 robot (Figure~\ref{fig:sim_results}). We compare our algorithm to three prior RL methods, TRPO~\cite{slmja-trpo-15}, RWR~\cite{ps-aenac-07}, and REPS~\cite{pma-reps-10}, as well as two prior GPS methods, including on-policy MDGPS~\cite{ml-gpsamd-16} and an older formulation based on the Bregman Alternating Direction Method of Multiplers (BADMM)~\cite{lfda-eetdv-16}. All methods use a neural network with two layers of 42 hidden units each, and results are averaged over 5 random seeds.

The planar reaching task involves a simulated 3 DoF arm attempting to reach a target position, with the joint angles, end-effector position, target position, and joint and end-effector velocity comprising the state. The state space has 18 dimensions, and the action space has 3, one torque control for each joint of the arm. The goal is to reach a target position drawn uniformly from a 2D region which nearly covers the reach of the arm. In our experiments, our method and prior deep RL methods begin with target positions drawn uniformly from this region, while prior GPS methods use 8 initial positions placed approximately in a circle, chosen to cover as much of the sampling region as possible.

The peg insertion task requires a simulated 7 DoF arm to insert a peg into a hole, using the joint angles, end-effector positions, and their time derivatives as the state representation. The end-effector positions are provided relative to the hole position, which tells the policy the location of the target. The state space has 26 dimensions, and the action space has 7. The goal is to learn a policy that can accurately insert the peg into the hole at various positions, distributed over either a 2D region of side length 0.4, or a 3D region of side length 0.3. In comparing to prior GPS methods that require deterministic resets to fixed initial states, we use initial positions at the corners of the range (4 for the 2D region, 8 for the 3D region). Our method and prior deep RL algorithms receive a state sampled uniformly at random from the training region for each sample.

In the experiments that combine perception and control in Section~\ref{sec:vision}, we repeat the tasks above using the spatial feature architecture presented in prior work~\cite{lfda-eetdv-16}. As shown in Figure~\ref{fig:vis_results}, this architecture takes downsampled RGB images as raw visual input, and uses convolutional layers to produce visual features. These features are then fed into a fully connected layers, along with joint angles/velocities (but not end effector or target positions). For the reacher task we use the same set-up as above, while for the peg insertion task we evaluate a 2D sampling region of side length 0.3.

\subsection{Generalization with Random Initial States}

Figure~\ref{fig:sim_results} shows the results for the reacher and the two peg insertion tasks, evaluated at test positions drawn uniformly from the sampling region, showing that our method is able to solve the reacher task in 600 episodes, the 2D region peg insertion task in 480 episodes, and the 3D region peg task in 800 episodes. To evaluate how well our method can handle random initial states, we compare to two prior GPS methods: MDGPS~\cite{ml-gpsamd-16} and BADMM-based GPS~\cite{lfda-eetdv-16}. The results indicate that the use of random initial states with trajectory-centric clustering still allows our method to effectively update the local policies. The improved generalization can also be explained due to the additional variability in states seen during training. We emphasize, however, that the main benefit of our method is not any improvement in generalization, but rather the ability to handle random initial states at all: in many cases, it is impossible or impractical to deterministically reset the system into a chosen initial state, in which case prior guided policy search methods cannot be used at all.

\subsection{Comparison to Prior Deep RL Methods}

Figure~\ref{fig:sim_results} also shows a comparison to a number of prior deep reinforcement learning methods that also do not require deterministic resets to consistent initial states. We used the rllab package~\cite{dchsa-bdrl-16} to compare to TRPO~\cite{slmja-trpo-15}, RWR~\cite{ps-aenac-07}, and REPS~\cite{pma-reps-10}. While we also attempted to compare to DDPG~\cite{slhdwr-dpga-16}, we were unable to find a hyperparameter setting that worked for our tasks. For each of the other methods, the hyperparameters were tuned to achieve the best possible sample efficiency. The results indicate that, although TRPO was able to eventually solve the task, all prior methods required substantially more samples, with TRPO requiring several orders of magnitude more data. This indicates that our approach maintains the sample efficiency of guided policy search methods while relaxing the consistent initial state assumption.

\begin{figure}[t]
    \centering
    \includegraphics[width=0.45\textwidth]{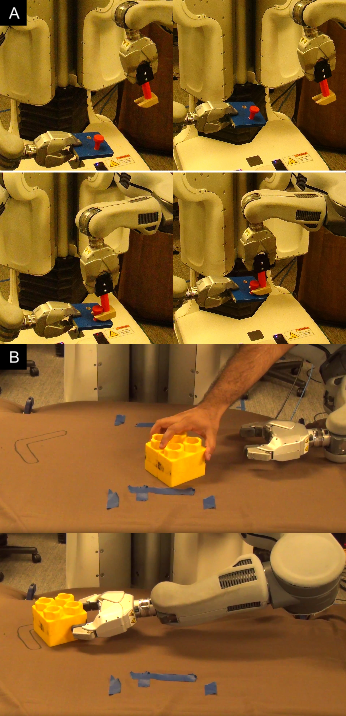}
    \vspace{-0.1in}
    \caption{PR2 tasks. For each task, top images show initial positions, while bottom images show successful final positions. (A) The PR2 must learn to accurately place the claw of the toy hammer underneath the nail, beginning from a random initial position within a $10~\text{cm} \times 10~\text{cm} \times 5~\text{cm}$ cuboid. (B) The PR2 must learn to push a randomly placed block to a target position using visual features.}
    \vspace{-0.1in}
    \label{fig:robot_tasks}
\end{figure}

\begin{figure*}[t]
    \centering
    \includegraphics[width=0.99\textwidth]{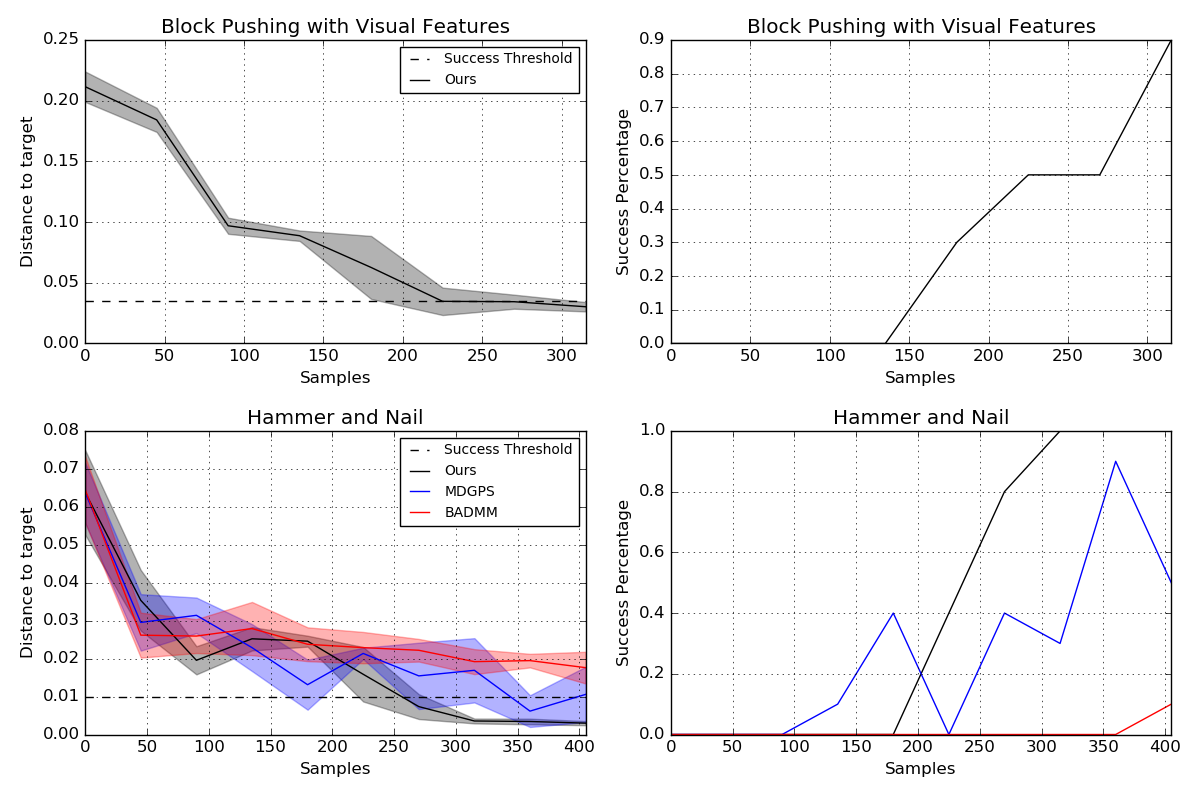}
    \vspace{-0.1in}
    \caption{Real-world robotic results. We compare our method to prior GPS methods on two real-world manipulation tasks using the PR2 robot. In the first task, the robot must place the claw of a toy hammer under a nail, and in the second task the robot must push a randomly placed block to a fixed target position. We plot the final distance to the target on test positions vs. the number of samples used during training, as well as the percentage of successful test runs, where success is defined by the dotted line in the left plot. Our method learns more quickly than prior GPS methods, and generalizes more effectively to test positions.}
    \vspace{-0.1in}
    \label{fig:robot_results}
\end{figure*}

\subsection{Learning with Vision}
\label{sec:vision}

Figure~\ref{fig:vis_results} shows results for our method when learning directly from simulated camera images on the simulated tasks. The reacher task required about 600-800 samples to obtain good results, while the peg insertion task required about 500. The results indicate that our method is able to train high dimensional policies represented by convolutional neural networks, which allows it to learn policies that combine both perception and control.

\subsection{Physical Experiments with the PR2 Robot}

In addition to the simulated experiments, we evaluated our method on two real-world tasks using a PR2 robot (Figure~\ref{fig:robot_tasks}). In the first task, the robot must learn to accurately fit the claw of a toy hammer under a nail, for a range of initial nail positions. For our method, the initial position of the nail is sampled randomly from a cuboid region $10~\text{cm} \times 10~\text{cm} \times 5~\text{cm}$ in size. For comparison to prior GPS methods, we use 9 fixed initial positions sampled from the same cuboidal region. As before, we emphasize that although we compared to prior GPS methods, these methods required the ability to deterministically reset the environment to one of the 9 initial positions, while our method did not. In order to make this comparison feasible, the robot held the target object (a board holding the nail) in its right gripper. However, in a real-world scenario, the nail might be attached to an arbitrarily positioned object in the world.

In the second task, the robot must learn to push a block to a target position. For our method, the block is placed randomly by the robot operator at the beginning of each trial, but we do not compare to prior GPS methods as they require deterministic resets. We do not compare to other prior deep RL methods on either task, as they required too many samples to be practical for real-world training.

As shown in Figure~\ref{fig:robot_results}, our method not only achieves a high success rate on this task, but is also able to generalize more widely despite using the same total number of samples, since the prior methods are only trained from a small number of initial positions. The real-world results therefore confirm our simulated findings: our proposed method can not only handle random initial states effectively, but in fact tends to produce faster learning and more generalizable policies. Video recordings of the robot learning process and the final policy can be found online.\footnote{\url{http://sites.google.com/site/resetfreegps}}

\section{Discussion and Future Work}

In this paper, we presented a new guided policy search method that lifts a major limitation of prior GPS approaches: the need to deterministically reset the environment into particular initial states in order to iteratively optimize each of the local policies. Our method introduces two key components that enable it to handle random initial states: first, we generate samples from the current global neural network policy, instead of the individual local policies, and second, we create new local policies dynamically at each iteration by clustering the rollouts. Our algorithm can train deep neural network policies with the efficiency of guided policy search, achieving sample complexities orders of magnitude lower than direct deep RL methods, while removing the deterministic reset requirement, allowing our GPS algorithm to be applied to general reinforcement learning problems with stochastic initial states. We experimentally demonstrate that randomly sampling initial states actually improves both generalization and learning speed over the standard guided policy search approach, and empirically compare our method to a number of prior RL algorithms.

Although our algorithm can accommodate random initial states, the use of time-varying linear-Gaussian dynamics models does limit its ability to handle arbitrary stochastic dynamics. Differentiable dynamical systems can always be approximated locally as time-varying linear systems, but the same does not necessarily hold for arbitrary stochastic systems, which might exhibit locally multimodal behavior, making the locally linear-Gaussian assumption problematic. However, Gaussian dynamics noise is a common and generally reasonable assumption for continuous physical systems, and we observed that our method was able to produce good results on a real physical robot. Nonetheless, properly accommodating non-Gaussian noise models is a promising and interesting direction for future work.

Our method lifts the deterministic reset requirement of GPS algorithms, but it still relies on the availability of a ``teacher'' algorithm to guide learning, which in our case is still the same model-based LQR approach used in prior work~\cite{la-lnnpg-14}. One promising direction for future work is to extend this approach to more powerful trajectory-centric RL methods, including stochastic optimization methods~\cite{tbs-rlmsh-10} and model-free trajectory optimization~\cite{aaan-mftor-16}. Another promising direction for future work is to explore further applications of sample-efficient deep reinforcement learning in stochastic environments. The experimental results presented in this work focus on robotic manipulation, but the same approach could be applied in domains ranging from autonomous flight to locomotion.


\bibliographystyle{IEEEtran}
\bibliography{references}

\end{document}